\documentclass[runningheads]{llncs}
\usepackage[T1]{fontenc}
\usepackage{graphicx}
\usepackage[nolist]{acronym}
\usepackage{amsmath}

\usepackage{algorithm}
\usepackage{algpseudocode}
\usepackage{amsmath,amssymb,amsfonts}
\usepackage{caption}
\usepackage{capt-of}
\usepackage{hyperref}
\usepackage[capitalise]{cleveref}
\usepackage{color}

\usepackage{longtable}
\usepackage{makecell}
\usepackage{mathtools}
\usepackage{multirow}
\usepackage{siunitx}
\usepackage{subcaption}
\DeclareSIUnit\pixel{px}
\usepackage{xspace}

\makeatletter
\DeclareRobustCommand\onedot{\futurelet\@let@token\@onedot}
\def\@onedot{\ifx\@let@token.\else.\null\fi\xspace}

\let\oldthefigure\thefigure
\let\oldthetable\thetable
\newcommand{\supplementaryfigures}{%
  \setcounter{figure}{0}
  \setcounter{table}{0}
  \renewcommand{\thefigure}{S\oldthefigure}
  \renewcommand{\thetable}{S\oldthetable}
}

\begin{document}
\newcommand{\mytitle}{Leveraging Image Captions for Selective Whole Slide Image Annotation}
\title{\mytitle}
\author{Jingna Qiu\inst{1} \and
Marc Aubreville\inst{2,3} \and
Frauke Wilm\inst{1,4} \and
Mathias Öttl\inst{1,4} \and
Jonas Utz\inst{1} \and
Maja Schlereth\inst{1} \and
Katharina Breininger\inst{1,5}
}
\authorrunning{J. Qiu et al.}
%
\institute{Department Artificial Intelligence in Biomedical Engineering, \\FAU Erlangen-Nürnberg, Erlangen, Germany
\email{jingna.qiu@fau.de}\\
\and Technische Hochschule Ingolstadt, Ingolstadt, Germany\\
\and Flensburg University of Applied Sciences, Flensburg, Germany\\
\and Pattern Recognition Lab, Department of Computer Science, \\ FAU Erlangen-Nürnberg, Erlangen, Germany\\
\and Center for AI and Data Science, Universität Würzburg, Würzburg, Germany\\}
\maketitle              
\begin{abstract}
Acquiring annotations for \acp{wsi}-based deep learning tasks, such as creating tissue segmentation masks or detecting mitotic figures, is a laborious process due to the extensive image size and the significant manual work involved in the annotation. This paper focuses on identifying and annotating specific image regions that optimize model training, given a limited annotation budget. While random sampling helps capture data variance by collecting annotation regions throughout the \ac{wsi}, insufficient data curation may result in an inadequate representation of minority classes. Recent studies proposed diversity sampling to select a set of regions that maximally represent unique characteristics of the \acp{wsi}. This is done by pretraining on unlabeled data through self-supervised learning and then clustering all regions in the latent space. However, establishing the optimal number of clusters can be difficult and not all clusters are task-relevant. This paper presents prototype sampling, a new method for annotation region selection. It discovers regions exhibiting typical characteristics of each task-specific class. The process entails recognizing class prototypes from extensive histopathology image-caption databases and detecting unlabeled image regions that resemble these prototypes. Our results show that prototype sampling is more effective than random and diversity sampling in identifying annotation regions with valuable training information, resulting in improved model performance in semantic segmentation and mitotic figure detection tasks. Code is available at \url{https://github.com/DeepMicroscopy/Prototype-sampling}.


\end{abstract}

\section{Introduction}
\acresetall
Deep learning models for \ac{wsi} semantic segmentation and mitotic figure detection can assist in tumor grading by identifying possible malignant areas~\cite{aubreville2020deep,bejnordi2017diagnostic,kim2021paip,veta2019predicting}. High-quality annotations are crucial to properly train these models, but the acquisition process of these annotations requires pathologists as annotators and is extremely time-consuming due to the large image size in the gigapixel range and the substantial manual effort involved in the annotation. Segmentation masks are generated at the pixel level, and careful examination of thousands of cells is necessary to exhaustively identify mitotic figures (cells undergoing division). This work focuses on discovering specific annotation regions that optimize model training, while leaving the rest of the image unlabeled. This can help reduce the need for manual labeling or accommodate a limited availability of expert annotators. Decreasing the annotation area has been shown to efficiently streamline the process of producing segmentation masks for \acp{wsi}~\cite{jin2021reducing,lai2021joint,qiu2023adaptive,yang2017suggestive}; but, to our knowledge, it has not been used for acquiring mitotic figure annotations. Here, annotators are frequently asked to exhaustively annotate manually selected regions of interest~\cite{aubreville2023mitosis}, which could be considerably simplified if cell examination was required only for a fraction of these regions that are identified as informative.

Given a pool of unlabeled \acp{wsi} and a certain annotation budget, the task is to find annotation regions that contain useful information for training the model in the downstream task. One natural idea is random sampling, which ensures data diversity by including tissue from different parts of the \ac{wsi}. However, it may raise issues such as the selected regions inadequately representing minority classes. Some works involve manually selecting annotation regions~\cite{jin2021reducing}, which necessitates expertise in estimating which information is most useful for model training. However, humans may not fully recognize the varying degrees of visual variety presented within each class~\cite{shin2021all}. In recent years, \ac{ssl} techniques have popularized diversity sampling, which identifies a set of regions that represent distinct features of the dataset with minimum duplication. Specifically, the pretrained model creates embeddings for all regions, the embeddings are clustered and annotation regions are selected at the center of each cluster~\cite{hacohen2022active,pourahmadi2021simple} or to maximally cover each cluster~\cite{jin2022one,zheng2019biomedical}. However, the optimal number of clusters is often not clear, and some clusters may feature irrelevant information for training the downstream task model, such as artifacts.

In this work, we introduce a novel method called \textit{prototype sampling} for the specified task. It identifies annotation regions that exhibit typical characteristics of each task-specific class without the need for clustering. Specifically, our method involves collecting prototype embeddings for each class from extensive histopathology image-caption databases, and identifying unlabeled regions that closely resemble these prototypes. Two databases are used in the study: \textit{ARCH}, composed of figures and captions from PubMed articles and textbooks~\cite{gamper2021multiple}, and \textit{OpenPath}, encompassing Tweets where pathologists discussed cases~\cite{huang2023leveraging}. These resources significantly decrease the time and challenge of searching for class prototypes. They offer a valuable and diverse collection of images showcasing both typical and ambiguous aspects of many diseases and include a wide range of data variations, including staining material and case origin. Similar to diversity sampling, our approach uses a pretrained model to extract feature embeddings for the following sampling process. However, we assign unlabeled regions to task-specific classes based on detected prototypes instead of relying on clustering results. One use case of our method is the construction of the initial labeled set of region-based \ac{al}~\cite{mackowiak2018cereals}, which iteratively trains the model on existing labeled regions and uses the trained model to determine which regions to annotate next. An initial labeled set that efficiently trains the first \ac{al} model can more reliably identify informative annotation regions in the subsequent \ac{al} cycle. Another application is to create a preliminary model by annotating a small portion of the data, and then use the model to automate further annotations or propose reference annotations~\cite{bertram2019large}. We test our method on semantic segmentation and mitotic figure detection tasks by training the model using the annotation regions selected through random, diversity, and prototype sampling.

\section{Method}
\subsection{General Setup}
Given a pool of $N$ unlabeled \acp{wsi} and an annotation budget of $n$ regions of size $l\times l$ per \ac{wsi}, the objective is to identify annotation regions containing useful training information for the downstream task. The annotation regions are detected based on a specified sampling method. In this section, we outline random sampling and diversity sampling, which serve as benchmark methods for our experiments. The random sampling method selects $n$ regions from each \ac{wsi} at random locations; regions with less than $10\%$ tissue area are excluded by tissue detection~\cite{otsu1979threshold}. We mainly followed~\cite{hacohen2022active} for diversity sampling. First, each \ac{wsi} is partitioned into a grid of regions of size $l\times l$. For each region we calculate the feature embedding using a pretrained model. We then aggregate the embeddings of all regions from $N$ \acp{wsi} and allocate them to $N*n$ clusters with K-means clustering, in order to prevent the selection of similar images regions from separate \acp{wsi}. Finally the regions centered at each cluster are selected. Next we describe prototype sampling in detail.

\begin{figure}[t]
\begin{minipage}[b]{\linewidth}
  \centering  \centerline{\includegraphics[width=\textwidth]{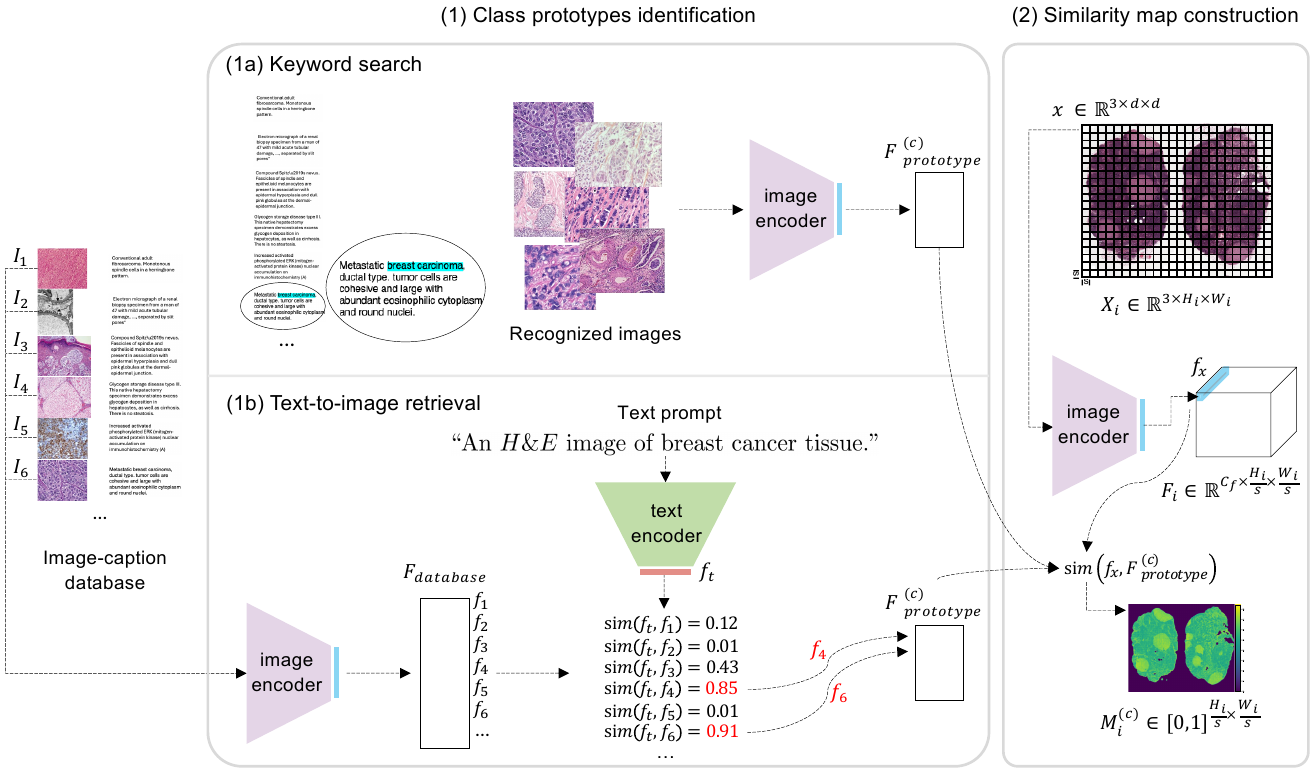}}
\end{minipage}
\caption{Workflow of prototype sampling.}
\label{fig:method}
\end{figure}
\subsection{Prototype Sampling}
Prototype sampling has two main components: 1) the acquisition of prototypes of each task-specific class from image-caption databases by keyword search or text-to-image retrieval, and 2) the construction of a similarity map that shows the similarity of different image areas to the prototypes. The similarity map is used for the identification of annotation regions. An illustration is shown in~\cref{fig:method}.

\noindent\textbf{Class Prototypes Identification}
\label{subsec:Class_Prototype_Images_Identification}
We obtain typical images of a specific class from histopathology image-caption databases by \textit{keyword search} or \textit{text-to-image retrieval}. \textit{Keyword search} exploits the detailed information contained in captions and is straightforward to conduct (\cref{fig:method} (1a)). Specifically, we create a set of ``with'' and ``without'' keywords, and recognize images with captions that include all ``with'' keywords and none of the ``without'' keywords. The ``with'' keywords are created using class names with several synonyms (e.g., ``cancer'' and ``tumor''), under the assumption that data analysis experts lack medical expertise. The ``without'' keywords are used to exclude irrelevant images, such as those of type photomicrograph. The recognized images are encoded into embeddings $\mathcal{F}_{prototype}^{(c)}$ using a pre-trained image encoder, with $c$ representing the class. \textit{Text-to-image retrieval} employs text and image encoders created through contrastive language-image pretraining (CLIP)~\cite{radford2021learning} to align visual concepts from images with the perception contained in the text (\cref{fig:method} (1b)). We use a text prompt ``An $H\&E$ image of $\{class\_name\}$ tissue.'' and obtain its embedding from the text encoder. All images in the database are encoded using the image encoder to create $\mathcal{F}_{database}$. $N_{prototype}$ database image embeddings that have the highest cosine similarity to the text prompt embedding are selected to form $\mathcal{F}_{prototype}^{(c)}$.

\noindent\textbf{Similarity Map Construction and Annotation Region Selection} 
We now identify image regions from the \ac{wsi} that are representative of the specified class. We partition the \ac{wsi} $X_{i}\in \mathbb{R}^{3\times H_{i}\times W_{i}}$ into patches with a stride of $s$ and then calculate their embeddings individually using a pretrained model. We denote a patch as $x\in \mathbb{R}^{3\times d\times d}$ and its embedding as $f_{x}\in \mathbb{R}^{C_f}$, where $C_f$ denotes the embedding dimension. The size of the patches $d$ and the level of magnification used to extract the patches from the \ac{wsi} pyramid are chosen according to the task. For instance, a small patch extracted at high magnification is preferable for comparing its resemblance to prototype images of mitotic figures, whereas patches characterizing a gland should be sufficiently large to encompass all tissue components. Note that each annotation region of size $l$ may contain multiple patches of size $d$. The representativeness of patch $x$ for class $c$ is defined as the highest cosine similarity between $f_{x}$ and the embeddings in $\mathcal{F}_{prototype}^{(c)}$, calculated as $sim(f_x, \mathcal{F}_{prototype}^{(c)})=max_{f_p\in \mathcal{F}_{prototype}^{(c)}} sim(f_x, f_p)$. Patch similarities are then combined into a similarity map $M_{i}\in [0, 1]^{\frac{H_{i}}{s}\times \frac{W_{i}}{s}}$ for the \ac{wsi}. The map is used to select annotation regions using a standard or an adaptive region selection method, denoted as ``prototype (standard)'' and ``prototype (adaptive)'' accordingly. The standard method~\cite{mackowiak2018cereals} moves a sliding window of the annotation region size (i.e., $\frac{l}{s}\times \frac{l}{s}$ (note that $l$ is defined at the original resolution)) across $M_{i}$ with a stride of $1$ pixel, evaluates the similarity of each region by summing up the similarities of patches within it, and then use non-maximum suppression to identify non-overlapping regions with the highest similarity values. The adaptive method~\cite{qiu2023adaptive} allows for selecting regions that can adjust in shape and size to accommodate variations in histopathological structures. The process includes the following steps: 1) Locate the pixel with the highest similarity $(u, v)$; 2) Create a binary mask by using a similarity threshold; 3) Identify the connected component that includes $(u, v)$ and choose its bounding box; and 4) Determine the similarity threshold through bisection search to ensure that the bounding box falls within the range $[\frac{1}{2}l\times \frac{1}{2}l, \frac{3}{2}l\times \frac{3}{2}l]$. 

\section{Experiments}
\subsection{Image-caption Pair Databases}
Two histopathology image-caption databases are used in the study. The ARCH database~\cite{gamper2021multiple} contains $8,617$ histology or immunohistochemistry image-caption pairs extracted from PubMed medical articles and $3,199$ pairs from 10 textbooks. The captions are manually curated to only include text related to diagnostic and morphological descriptions. The OpenPath database~\cite{huang2023leveraging} consists of $116,504$ image-caption pairs from Twitter posts and $59,869$ pairs from replies, gathered from 32 pathology subspecialty-specific hashtags. The database does not contain the extracted images and captions but only the links to the original posts. Image and caption embeddings obtained from PLIP~\cite{huang2023leveraging}, which is trained by fine-tuning a pre-trained CLIP model~\cite{radford2021learning} on OpenPath, are included in the database. 

\subsection{Semantic Segmentation}
\noindent\textbf{Dataset}
The public dataset CAMELYON16~\cite{litjens20181399} consists of $399$ \acp{wsi} of \ac{he}-stained excised regional sentinel lymph nodes. The task is the detection of the presence and extent of breast cancer metastases. Each \ac{wsi} with metastases is accompanied by a segmentation mask that outlines all metastases at the pixel level. We followed data usages in~\cite{qiu2023adaptive} for fair comparisons.

\noindent
\textbf{Training Setups}
We followed~\cite{qiu2023adaptive} using a segmentation framework based on patch classification, including all training setups, except for replacing the MobileNet~\cite{sandler2018mobilenetv2} encoder with the two following encoders: 1)\textit{ResNet18\_SSL}~\cite{ciga2022self}: A ResNet18 that is self-supervised pretrained on $0.4$ million histological patches with SimCLR~\cite{chen2020simple}; 2) \textit{ViT\_PLIP}: The image encoder (ViT-32-Base) of PLIP. Training and inference patches were extracted at the resolution of $0.25$~\si{\micro\meter\per\pixel} with $256 \times 256$ \si{pixels} and $224 \times 224$ \si{pixels} for the two encoders, respectively.

\noindent
\textbf{Creating Class Prototypes}
We used the ARCH and OpenPath databases for obtaining class prototypes for the experiments with the two encoders. From ARCH, we extracted breast cancer images by keyword search. We used \{breast, abnormal (tumor, cancer, carcinoma, metastases, metastasis, metastic)\} and \{IHC (immunohistochemical, immunohistochemistry, immunostain), photomicrograph (photomicrography)\} as the ``with'' and ``without'' keywords, respectively, where synonyms are given in the parentheses. In total $21$ images were found. Details on the pre- and post-processing steps can be found in~\cref{tab:pre_and_post_ARCH_keyword_search}. These images were processed by the \textit{ResNet\_SSL} encoder to generate image embeddings. Prior to this, the images were center-cropped to the size of model input to maintain the original magnification and prevent distortion of the structures. From the OpenPath database, where the image embeddings are provided, we selected the top $N_{prototype}=100$ image embeddings under the hashtag ``\#BreastPath'' that best align with the embedding of the context prompt of ``An \textit{H}\&\textit{E} image of breast tumor tissue.'' generated by the PILP text encoder.

\noindent
\textbf{\Ac{wsi} Similarity Map Calculation}
Patches of size $256 \times 256$ \si{pixels} were extracted at resolution $0.25$~\si{\micro\meter\per\pixel} with a stride of $256$ \si{pixels} to ensure the incorporation of details required for metastases identification.

\noindent
\textbf{Comparison Methods} 
We compared to random sampling and diversity sampling. For fair comparisons, the two encoders were used to obtain region features in diversity sampling instead of pretraining a new model on the unlabeled data. Prior to this, the regions were downsampled to match the model input sizes.

\noindent
\textbf{Evaluation Scenarios and Metrics}
Recent works have found that the efficiency of a sampling strategy may differ depending on the the number and size of the annotation regions ~\cite{luth2023navigating,mackowiak2018cereals,qiu2023adaptive}, we follow~\cite{qiu2023adaptive} to evaluate on nine hyperparameter settings by combining $n\in\{1, 3, 5\}$ and $l\in\{4096,\,8192,\,12288\}$ \si{pixels}. We denote these three sizes as S (small), M (medium) and L (large), respectively. Slide-averaged intersection over union across test slides containing tumor (mIoU (tumor)) was used as the evaluation metric, following~\cite{qiu2023adaptive}.

\subsection{Mitotic Figure Detection}
\noindent\textbf{Dataset}
The MITOS\_WSI\_CMC~\cite{aubreville2020completely} dataset consists of $21$ slides of canine mammary carcinoma that are fully annotated with $13,907$ mitotic figures. The task is to identify all mitotic figures in the \acp{wsi}, as mitotic count is a key criteria for tumor grading~\cite{cw1991value}. We followed data usages in~\cite{aubreville2020completely} for fair comparison.

\noindent\textbf{Training Setups}
We followed~\cite{aubreville2020completely} using a RetinaNet~\cite{lin2017focal} (ResNet18 backbone pretrained on ImageNet) as the mitotic figure detection model and the same training schemes, the only modification was sampling $1,000$ training patches per epoch instead of $5,000$ when the annotated area was $<5\%$ to avoid overfitting.

\noindent\textbf{Creating Class Prototypes}
We used ARCH to extract prototypes of mitotic figures by keyword search. The ``with'' keyword set included \{arrow (arrowhead, circle), mitotic (mitoses)\}, as mitotic figures are often highlighted with arrows or circles (see examples in~\cref{fig:qualitative_MITOS} (a-b)). 
In total $19$ images were identified, from each a section including the mitotic figure was manually cropped. We resized all sections to $64 \times 64$ \si{pixels} and calculated features using the ResNet18 backbone. 

\noindent\textbf{\Ac{wsi} Similarity Map Calculation}
Patches of size $64 \times 64$ \si{pixels} ($0.25$~\si{\micro\meter\per\pixel}, a stride of $64$ \si{pixels}) were extracted to include only one or a few cells per patch.

\noindent\textbf{Comparison Methods} 
The same settings in the CAMELYON16 experiments were used, using the ResNet18 backbone to extract region features.

\noindent\textbf{Evaluation Scenarios and Metrics}
The same settings in the CAMELYON16 experiments were used. The three region sizes $l\in\{4096,\,8192,\,12288\}$ \si{pixels} correspond to annotation areas of $\{1.05,\,4.19,\,9.44\}$ \SI{}{\milli\metre\squared}, enclosing the recommended area for performing mitotic count ($10$ high power field (HPF), $2.37$ \SI{}{\milli\metre\squared}). The F1 score was used as the evaluation metric, following~\cite{aubreville2020completely}.

\subsection{Results}
\begin{figure}[t]
\begin{minipage}[b]{\linewidth}
  \centering  \centerline{\includegraphics[width=0.9\textwidth]{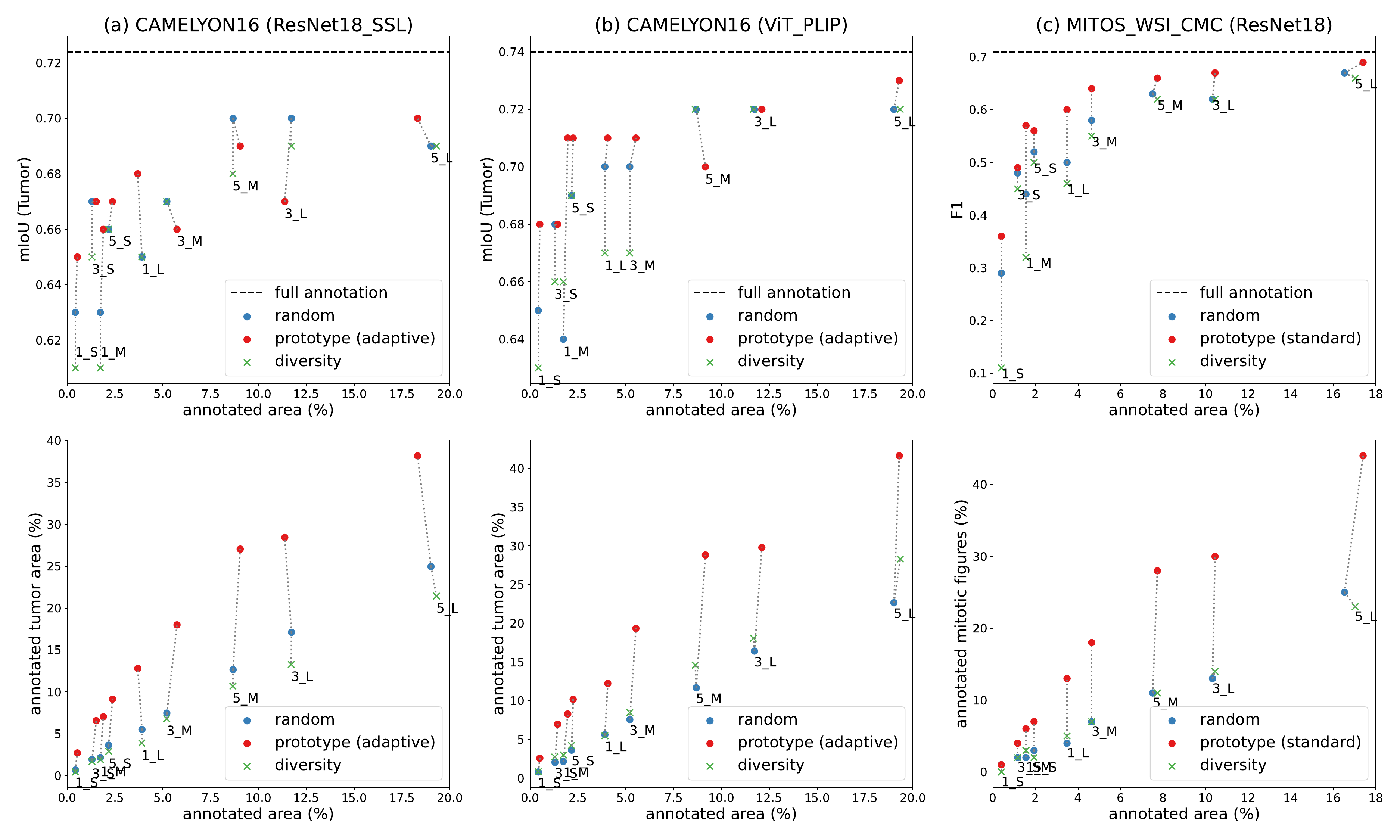}}
\end{minipage}
\caption{(a-b) Results on CAMELYON16 dataset: mIoU (Tumor) and annotated tumor area (\%) as functions of annotated tissue area (\%). (c) Results on MITOS\_WSI\_CMC dataset: F1 and the ratio of annotated mitotic figures as functions of annotated tissue area (\%). Prototype (adapt) can have different amounts of annotated area as the size of each selected region is dynamically determined. All other methods select regions of size $l\times l$. Random sampling can lead to a slightly smaller annotated area when no more non-overlapping region containing at least $10\%$ tissue is found. All results show median values from five repetitions.}
\label{fig:quantitative}
\end{figure}

\begin{figure}[t]
\begin{minipage}[b]{\linewidth}
  \centering  \centerline{\includegraphics[width=\textwidth]{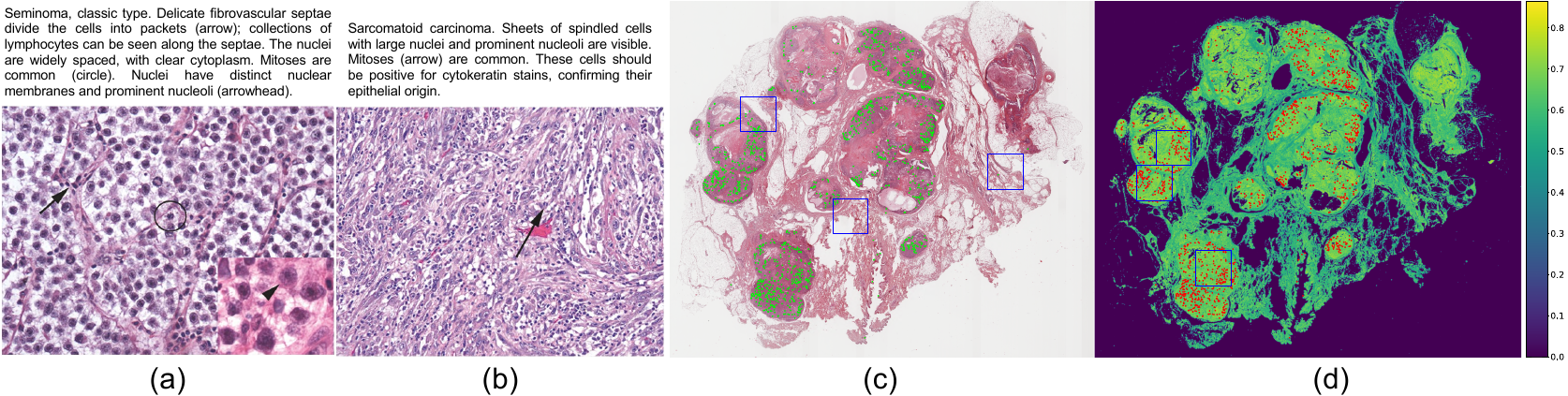}}
\end{minipage}
\caption{(a-b) Two example prototype images of mitotic figure. (c-d) An example \ac{wsi} and its similarity map. Ground truth mitotic figures marked green in (c) and red in (d). The blue boxes in (c) and (d) indicate regions selected with random and prototype sampling, respectively (hyperparameter: 3\_M).}
\label{fig:qualitative_MITOS}
\end{figure}

\begin{figure}
\begin{minipage}[b]{0.9\linewidth}
  \centering  \centerline{\includegraphics[width=\textwidth]{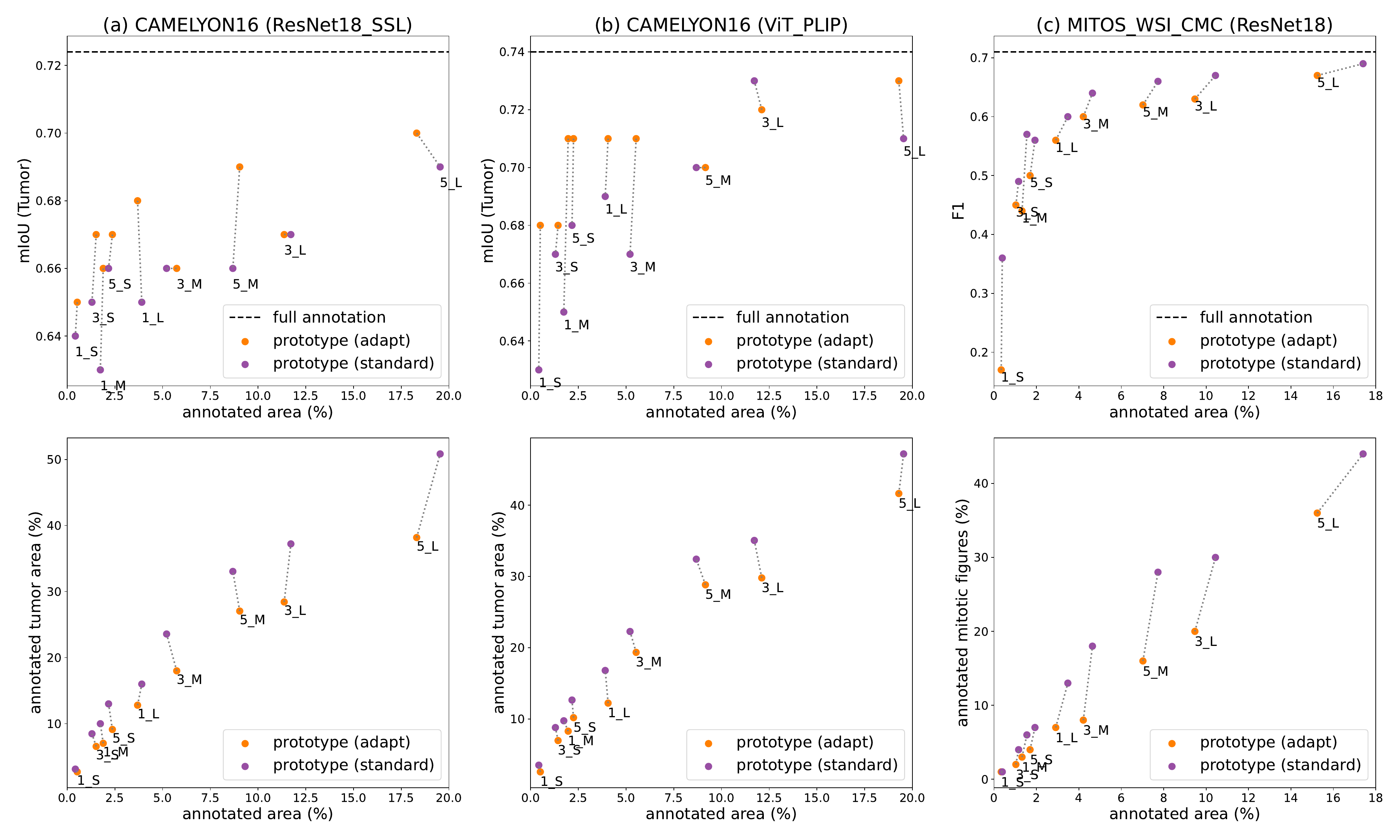}}
\end{minipage}
\caption{Comparison of the standard and adaptive region selection method. (a-b) Results on CAMELYON16 dataset: mIoU (Tumor) and annotated tumor area (\%) as functions of annotated tissue area (\%) for prototype sampling across various hyperparameter settings. (c) Results on MITOS\_WSI\_CMC dataset: F1 and the ratio of annotated mitotic figures as functions of annotated tissue area (\%) for prototype sampling across various hyperparameter settings. All results show the median value from five repetitions.}
\label{fig:adaptive_standard}
\end{figure}

\Cref{fig:quantitative} shows the results across all methods. Given a certain budget for annotation area, prototype sampling is more effective than random and diversity sampling in creating a labeled set that maximizes model outcome on both tasks of breast cancer metastases segmentation and mitotic figure detection. This could be attributed to the increased ratios of tumor area and mitotic figures identified and annotated. Notably, prototype sampling achieves near full annotation performance in CAMELYON16 (ViT\_PLIP) and MITOS\_WSI\_CMS (ResNet18) experiments by annotating less than $20\%$ of the tissue area, without the iterative process of annotation region identification in the conventional \ac{al} procedures, leading to substantially reduced computational costs.
Specifically, on the CAMELYON16 dataset, prototype sampling shows a smaller performance benefit over the other two methods when selecting a large number of regions per \ac{wsi}, (e.g., in the cases of 3\_L and 5\_M). This might be because multiple similar regions  all possessing tumor characteristics are selected from the same \ac{wsi}, leading to low data diversity. A comparison between the adaptive and standard region selection methods in \cref{fig:adaptive_standard} provides further evidence to support this hypothesis: Annotation regions selected using ``prototype (standard)'' contain a larger amount of tumor area than the annotation regions selected by ``prototype (adapt)'', but the trained model shows inferior performance in segmenting tumors on the test set. The standard method can identify a larger area of tumor for annotation as it selects regions that have the highest number of patches resembling tumor, while the adaptive method selects annotation regions centered at the patches that are most similar to tumor prototypes.
On the MITOS\_WSI\_CMC dataset, prototype sampling identifies annotation regions containing more mitotic figures than the other two methods, leading to a consistent increase in model performance. Here, ``prototype (standard)'' also outperforms ``prototype (adapt)'' as regions with a larger number of patches resembling mitotic figure are identified for annotation (\cref{fig:adaptive_standard}). \Cref{fig:qualitative_MITOS} (d) provides an example of similarity map, where areas with patches that closely resemble mitotic figure prototypes align well with the areas containing ground truth mitotic figure annotations. 

\section{Discussion and Conclusion}
We presented a novel method for selecting valuable annotation regions that allows to effectively train the model without annotating the entire \acp{wsi}. These regions are detected as possessing typical characteristics of each task-specific class based on their strong resemblance to prototypes recognized from image-caption databases by keyword search or text-to-image retrieval. The efficacy of our method has been proven in different tasks and evaluation settings. 

Experiments on the breast cancer metastases segmentation task using both ARCH and OpenPath databases allow evaluating the impact of image-caption database quality on method robustness. While ARCH is well-curated with manually selected samples and cleaned captions, OpenPath is only subjected to image quality sampling testing and automatic text cleaning. We expected that prototypes obtained from OpenPath may have more noise, but we see no drawback of using OpenPath in our experiments. This evaluation was prioritized over an examination involving artificially introduced noise patterns (e.g., swapping captions of two randomly selected pairs), as it depicts ``real-world'' database quality fluctuations. Future work will investigate the impact of prototype set size.

Particularly, our method can aid the annotation of minority classes. While these classes may also have limited presence in image-caption databases, our method help identify rare but existing samples or discover prototypes of related diseases using prior knowledge. Concerns about finding prototypes for rare diseases are further alleviated by ongoing efforts in creating larger and more diverse histopathology image-caption databases to build powerful all-purpose models.

\begin{acronym}
\acro{wsi}[WSI]{whole slide image}
\acro{he}[H\&E]{Hematoxylin \& Eosin}
\acro{froc}[FROC]{Free Response Operating Characteristic}
\acro{iou}[IoU]{intersection over union}
\acro{miou}[mIoU]{mean intersection over union}
\acro{al}[AL]{active learning}
\acro{ssl}[SSL]{self-supervised learning}
\acro{roi}[RoI]{region of interest}
\end{acronym}

\begin{credits}
\subsubsection{\ackname} We acknowledge support by d.hip campus - Bavarian aim (J.Q. and K.B.), the German Research Foundation (DFG) project 460333672 CRC1540 EBM, project 405969122 FOR2886 Pandora, as well as the scientific support and HPC resources provided by the Erlangen National High Performance Computing Center (NHR@FAU) of the Friedrich-Alexander-Universität Erlangen-Nürnberg (FAU). The hardware is funded by DFG.

\subsubsection{\discintname}
The authors have no competing interests to declare that are relevant to the content of this article.
\end{credits}

%
%
\bibliographystyle{splncs04}
\bibliography{Paper-2268}
\newpage 
\setcounter{page}{1}
\setcounter{section}{0}
\title{\mytitle}
\author{Jingna Qiu\inst{1} \and
Marc Aubreville\inst{2,3} \and
Frauke Wilm\inst{4} \and
Mathias Öttl\inst{4} \and
Jonas Utz\inst{1} \and
Maja Schlereth\inst{1} \and
Katharina Breininger\inst{1,5}
}
\authorrunning{Qiu et al.}
%
\institute{Department Artificial Intelligence in Biomedical Engineering, \\FAU Erlangen-Nürnberg, Erlangen, Germany
\email{jingna.qiu@fau.de}\\
\and Technische Hochschule Ingolstadt, Ingolstadt, Germany\\
\and Flensburg University of Applied Sciences, Flensburg, Germany\\
\and Pattern Recognition Lab, Department of Computer Science, \\FAU Erlangen-Nürnberg, Erlangen, Germany\\
\and Center for AI and Data Science, Universität Würzburg, Würzburg, Germany}
\maketitle              

\section{Supplementary Materials}
\supplementaryfigures

\begin{table}
\caption{Pre- and post-processing steps for obtaining class prototype images by keyword search from the ARCH database.}\label{tab:pre_and_post_ARCH_keyword_search}
\begin{tabular}{|m{1.8cm}|m{5.1cm}|m{5.1cm}|}
\hline
 &  Method & Examples\\
\hline
Pre-processing & 1. Subcaptions division by pattern recognition (the subfigures have been individually split by the data supplier, each accompanied by the original caption). Before image search, we divide subcaptions by recognizing patterns.  2. Any supcaption that is applied to all subfigures are appended to each subcaption. & We used the following patterns: ``A, xxx.'', ``(A) xxx.'', and ``xxx (A).'', where A and xxx represent the subfigure identifier and the subcaption.\\
\hline
Post-processing &  Manually clean the recognized images. & We removed an image with caption containing ``melanoma metastatic to the breast'', it was recognized as containing keywords ``breast'' and ``metastatic''.
We removed an image with caption containing `lack mitotic activity'', it was recognized as containing keywords `mitotic''.\\
\hline
\end{tabular}
\end{table}

\begin{table}
\caption{Our method uses class names as keywords for prototype image retrieval, under the assumption that data analysis experts lack medical expertise. This design may lead to an incomplete retrieval of prototype images. We share several pitfalls we noticed during experiments and our solutions/suggestions for promoting future investigations.}\label{tab:pitfalls}
\begin{tabular}{|m{4cm}|m{4cm}|m{4cm}|}
\hline
Problem &  Example & Solution/Suggestion \\
\hline
The class name is not included in the caption. & The phrase ``SNL micrometastasis possibly derived from breast metastasis'' cannot be located using the search term ``breast cancer''.  & We alleviated this type of problems by offering synonyms.\\
\hline
Complex medical terminology is employed to describe the condition without explicitly stating the name of the disease. &  ``Ductal carcinoma in situ'' refers to a specific type of breast cancer.  & Having a list of medical terms offered by specialists is beneficial.\\
\hline
Contextual information in the caption hinders accurate image recognition. & A huge nodal melanocytic nevus image was detected while searching for breast cancer images because the caption contains ``Incidental large nevus found during an axillary node dissection in a patient with breast carcinoma.''. & We resolved this type of error by manually cleaning irrelevant images.\\
\hline
\end{tabular}
\end{table}
\end{document}